# Intrinsically Motivated Learning of Visual Motion Perception and Smooth Pursuit

Chong ZHANG, Yu ZHAO, Jochen TRIESCH and Bertram E. SHI

*Abstract—* We extend the framework of efficient coding, which has been used to model the development of sensory processing in isolation, to model the development of the perception/action cycle. Our extension combines sparse coding and reinforcement learning so that sensory processing and behavior co-develop to optimize a shared intrinsic motivational signal: the fidelity of the neural encoding of the sensory input under resource constraints. Applying this framework to a model system consisting of an active eye behaving in a time varying environment, we find that this generic principle leads to the simultaneous development of both smooth pursuit behavior and model neurons whose properties are similar to those of primary visual cortical neurons selective for different directions of visual motion. We suggest that this general principle may form the basis for a unified and integrated explanation of many perception/action loops.

## I. INTRODUCTION

The efficient encoding hypothesis posits that neurons respond so that they best represent sensory data while requiring as few neurons to respond as possible [1]. Mathematically, this concept has been captured by sparse coding algorithms, which seek to represent input vector vectors as linear combinations of basis functions drawn from a possibly overcomplete dictionary. These representations are sparse in the sense that only a few of the basis functions are used, i.e. most of the coefficients of the linear combination are zero. It has been demonstrated that basis functions from dictionaries learned to best represent natural input stimuli closely resemble the receptive fields of neurons found in the primary sensory cortices for vision [2] and audition [3]. The concept of efficient encoding is attractive as a model for biological neural systems, since neural firing is metabolically expensive.

A key prediction of the efficient encoding hypothesis and sparse coding models is that the sensory processing in the brain reflects the statistics of the sensory input. An important question then is what determines these statistics. The composition of the natural environment is certainly a key determinant of the statistics. Indeed, this has driven much work in using natural images and natural sounds in sparse coding algorithms. However, this picture is incomplete without accounting for the equally important effect of behavior, which shapes the statistics by directing the sensory apparatus preferentially towards some regions of the environment. Accounting for the effects of behavior is complex, since the behavior itself is often driven by the same sensory neurons that develop. Thus, it is highly likely that behavior and sensory processing co-develop in an organism. To date, this problem of co-development has received little attention.

In this paper, we address the problem of co-development of behavior and perception by extending the efficient encoding hypothesis to include the effect of behavior. Our model posits that not only does sensory processing develop to best represent its input under constraints of limited resources (neural firing), behavior develops simultaneously to shape the statistics of the input so that it easier to encode. Both perception and behavior develop so as to maximize the faithfulness of the sensory representation, or equivalently, minimize reconstruction error. This developmental model is intrinsically motivated, since the reconstruction error is generated within the agent, and is determined by the fidelity of the agents internal representation of the environment.

We study this problem specifically in the context of visual motion perception and smooth pursuit behavior. This is an ideal test bed for this model, since evidence indicates that motion perception and smooth pursuit co-develop. The neural signal controlling pursuit is closely related to motion perception [4]. Smooth pursuit in infants does not appear until about 2 months of age [4]. It becomes more prominent and refined with age, concurrently with the development of more advanced motion extrapolation mechanisms [5]. Our experiments reported here indicate that both smooth pursuit and motion perception can develop as emergent properties of our framework.

## II. METHODS

The embodiment of our framework for smooth pursuit eye movements is illustrated in Figure 1. An eye which can rotate in both pan and tilt senses a time varying pattern of light projected onto its retina from the environment. The environment contains targets moving in the environment at random speeds, where the targets and speeds are changed randomly at periodic intervals.

The time varying retinal images from a foveal region are passed to a sparse coding stage, which seeks to represent local patches from the fovea by sparse linear combinations of spatio-temporal basis functions drawn from an overcomplete

This work was supported in part by the Hong Kong Research Grants Council (Program 619111) and the German DAAD (grant G HK25/10), the European Community (grant FP7-ICT-IP-231722) and the German BMBF (grant FKZ 01GQ0840).

Chong ZHANG, Yu ZHAO and Bertram E.SHI are all with the Department of Electrical and Computer Engineering, Hong Kong University of Science and Technology, Clear Water Bay, Kowloon Kong Hong (email: eebert@ust.hk). Bertram E. SHI is also with the Division Biomedical Engineering at HKUST

Jochen Triesch is with the Frankfurt Institute for Advanced Studies in Frankfurt am Main, Germany.

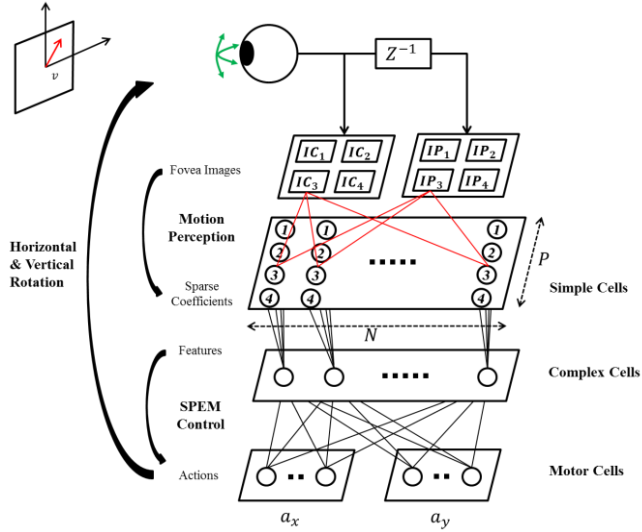

Figure 1 The developmental model for smooth pursuit. $P$ is the number of patches we extract in the fovea region, $N$ is the size of the dictionary, $a_x$ and $a_y$ are actions, $IP$ and $IC$ are fovea images obtained at previous and current time. For clarity we show four patches, but our systems uses 100.

dictionary. In a biological neural system, a possible substrate for this computation is a set of primary visual cortical neurons, where each basis function determines the spatio-temporal receptive field of one neuron [8], and the neuron's response at time $t$ is the coefficient of the corresponding basis function in the sparse representation of the incoming data at time $t$.

The model simple cells' responses are then squared and pooled spatially across simple cells in different patches with similar selectivity, similar to the way simple cell responses are combined in energy models of primary visual cortical cells tuned to orientation, disparity and motion. The complex cell responses are then fed to neural networks which determine the probability of different actions which may be taken to alter the rotational velocities of the eye.

The basis functions of the sparse coding stage and the weights of the action neural networks are initialized randomly. Thus, sensory processing and behavior at the start of development are independent of the environment. The basis functions and weights co-evolve to maximize the same criterion: fidelity of the linear combination in the sparse coding stage. As described in more detail below, basis functions are updated similarly to past work in sparse coding. The weights of the actor neural network are updated using the natural actor-critic reinforcement learning algorithm, where the reward to be maximized is the discounted sum of the negative of the current and future reconstruction errors.

We emphasize that the criterion optimized in the model is generic, and is not specific to smooth pursuit. Nonetheless, our experimental results in Section 3 will demonstrate (1) that the system develops smooth pursuit tracking behavior and (2) that the basis functions that develop have properties similar to those associated with the linear spatio-temporal receptive field profiles of motion tuned neurons in the primary visual cortex. In addition, because the smooth pursuit behavior shapes the statistics of the retinal slip (the difference between the projected target velocity and the eye rotational velocity) towards low retinal image velocities, the distribution of velocity tunings in the basis functions is biased towards lower retinal image velocities than would be presented to a stationary eye in the same environment. In the discussion, we suggest that this finding may provide a developmental explanation for the presence of a perceptual bias towards lower image velocities that has been found in psychophysical experiments.

The following describes the individual components of this model in more detail.

*A. Sparse coding*

In the model, we discretize time so that foveal images are presented at sampling rate of 30 frames per second. We assume that the foveal region covers 11 degrees of visual angle, and is sampled spatially at 5 pixels per degree. Thus, the dimension of each foveal image is 55 by 55 pixels. Each fovea image is further divided into $P = 100$ 10 by 10 pixel patches, each covering two degrees of visual angle (10 by 10 pixels). The patches cover the whole fovea image with 1 degree (5 pixel) overlap between neighboring patches both horizontally and vertically.

Sparse coding is applied to each patch independently, but all patches share the same dictionary, in the same way that neighboring image patches in a convolutional neural network share the same weights. For each patch, the sparse coding seeks to jointly encode information from both the current and previous time step. Thus, the input vector is 200 dimensional (10 by 10 pixels by 2 frames). We combine information for two frames because V1 evidence from [15] suggests that input to the direction selective cells in V1 modeled in our work comes from two distinct sub populations of cells whose temporal peak responses (68 and 93ms) differ by 25ms, approximately to one frame in our simulations.

We denote the input vector of image intensities at time t from patch $i \in \{1,...,P\}$ by $x_i(t)$. We reconstruct each input as a weighted sum of unit norm basis functions taken from an over-complete dictionary $\phi_n(t)$, where $n \in \{1,...,N\}$ indexes the basis functions and $N = 300$ is the size of the dictionary. The approximation is given by

$$x_i(t) \approx \sum_{n=1}^{N} a_{i,n}(t) \phi_n(t) \quad (1)$$

We use the matching pursuit algorithm proposed by Mallat and Zhang [11] to choose the coefficients $a_{i,n}(t)$ such that at most 10 of them are nonzero. We update the dictionary of bases using an on-line two step procedure similar to that used by Olshausen [2]. In the first step, we find the coefficients $a_{i,n}(t)$ using matching pursuit. In the second step, we assume the coefficients $a_{i,n}(t)$ are fixed, and adapt the bases to minimize the average normalized squared reconstruction error over patches

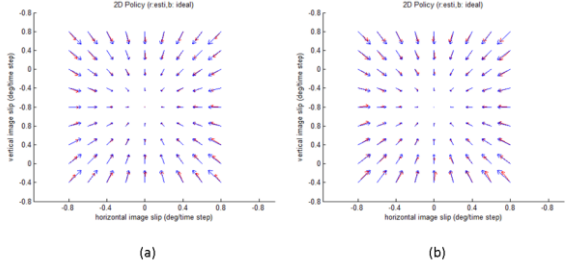

Figure 2 The policy obtained at the end of training for the Gaussian (a) and softmax (b) policy networks. The base of each arrow indicates the retinal slip. The direction and magnitude of each arrow shows the change in image slip after taking changing the eye's rotational velocity according to the policy assuming the target velocity remains constant. Blue arrows indicate an ideal policy, which would zero out retinal slip in one time step. Arrows point directly towards the origin with length proportional to the magnitude of image slip. Red arrows indicate the effect of the average command taken by the final policy. See text for details.

$$r(t) = \frac{1}{P}\sum_{i=1}^{P}\frac{\left\|x_i(t) - \sum_{n=1}^{N} a_{i,n}(t)\phi_n(t)\right\|^2}{\left\|x_i(t)\right\|^2} \quad (2)$$

Since all patches share the same dictionary, basis function updates are first pooled over patches before updating the dictionary. After each update, the bases are re-normalized so that they are unit norm.

Model complex cell responses pool the squared coefficients for each basis function over the set of all patches. Each basis function yields one complex cell response, resulting in a feature vector

$$\mathbf{f}(t) = \begin{bmatrix} f_1(t) & f_2(t) & \ldots & f_N(t) \end{bmatrix}^T \quad (3)$$

where

$$f_n(t) = P^{-1}\sum_{i=1}^{P} a_{i,n}(t)^2 \quad (4)$$

If we consider the coefficients $a_{i,n}(t)$ as models of simple cell output, this combination is consistent with complex cell model then.

*B. Reinforcement learning*

The weights in the action network evolve according to the natural actor critic reinforcement learning algorithm [12]. Reinforcement learning algorithms seek a policy mapping the current state of the agent to a probability distribution over potential actions by the agent, such that the discounted sum of future rewards is maximized. In our framework, the state of the agent is represented by the complex cell feature vector $\mathbf{f}(t)$. The actions are accelerations applied to modify the rotational velocities of the eye in the pan and tilt directions. The instantaneous reward at time $t$ is the negative of the squared reconstruction error (2). The discount factor is 0.3.

We use neural networks to represent the policy (the actor) and the value function (the critic). The input to the neural networks are the complex cell feature vector $\mathbf{f}(t)$. We compared the use of two different policy parameterizations: a softmax policy and a Gaussian policy.

For the softmax policy, actions are chosen from a discrete set of $K$ possible actions whose probabilities are given by the outputs of a linear network with a softmax output nonlinearity, where the number of outputs is $K$. Since separate actions update the rotational velocities of the pan and tilt axes of the eye, we use two separate neural networks for each axis. Denoting the probability of choosing the $i^{th}$ action at time $t$ by $\pi_i(t)$,

$$\pi_i(t) = \frac{\exp(z_i(t)/T)}{\sum_{j=1}^{K}\exp(z_j(t)/T)} \quad (5)$$

where the temperature $T$ is a positive scalar controlling the greediness of the policy, and $z_i(t)$ is the activation of the $i^{th}$ output neuron, which is computed by

$$z_i(t) = \mathbf{\theta}_i^T(t)\mathbf{f}(t) \quad (6)$$

where $\mathbf{\theta}_i(t)$ denotes the weights connecting to the $i^{th}$ neuron.

For the Gaussian policy, continuous-valued actions are chosen by sampling from a Gaussian distribution whose mean varies with $\mathbf{f}(t)$ but whose variance is fixed. The mapping from $\mathbf{f}(t)$ to a mean vector is performed by a three layer neural network, with $N$ neurons in the input layer, $H = 5$ neurons in the hidden layer whose activation functions are hyperbolic tangents, and two neurons in the output layer, which is linear. The two neurons encode the means of 1D Gaussian distributions describing the distributions of the pan and tilt acceleration actions.

*C. Environment model*

For simplicity in the simulations, we assume that the targets are textures applied to a sphere centered at the center of the eye and extend to fill the fovea. For textures, we use 20 images taken from van Hateren database [10]. Every 1/3 of a second (10 frames), we change the target texture and choose a new rotational velocity for the target. The rotational velocity is kept constant until the next change. We constrained the target rotational velocity to be at most 24 deg/s in the pan and tilt directions. The rotational velocity of the eye was constrained to the same range, which is consistent with the maximum speed of around 30 deg/s for smooth pursuit directions in humans. The maximum retinal slip is thus 48 deg/s in each dimension, if the eye and target move in opposite directions. Using our discretization, this corresponds to a range of horizontal and vertical velocities ranging between +8 and -8 pixels per frame.

The eye's rotational velocity at time t is obtained by taking the rotational velocity at time t-1 and adding the acceleration action obtained by sampling from the policy distribution. For the softmax policies, the network encodes K = 11 commands, which are equally spaced accelerations ranging from -900 to +900 deg/s2 (-5 to +5 pixels/frame2)

## III. RESULTS

We simulated the model using both the softmax and Gaussian policies starting from random initial conditions. For each case, we conducted three trials. Both models led to the emergence of smooth pursuit in all trials, indicating that the exact structure of the model is not critical for the emergence of this behavior. In the beginning of the simulations, the rotational velocity of the eye is essentially random, but the eye begins to follow the target as both basis functions and the policy network develop.

Figure 2 shows the effect on retinal slip in the next time step for the average command taken by the final policy for different initial values of retinal slip. This figure illustrates clearly the emergence of a smooth pursuit behavior, since the average actions taken by both the softmax and Gaussian policies are quite close to those of an "ideal" smooth pursuit policy that would zero out retinal slip in one time step. The range of slips tested ranged between -0.8 to 0.8 deg/frame (-24 to 24 deg/s) in each direction in steps of 0.2 deg/frame (6 deg/s). For each value of retinal slip, the output of the policy network may vary with the foveal image content. To reduce the effect of this input driven variability, we average the greedy (maximum likelihood) action of the over 50 pairs of current and past foveal images with the same retinal slip. For the Gaussian policy, the greedy action is determined by the means of the Gaussians for pan and tilt. For the softmax policy, the greedy action corresponds to the maximum output of the softmax network. The foveal images were obtained by taking pairs of 55 by 55 pixel subwindows from images in the van Hataren database not used during training, where the locations of the subwindows differ by a translation corresponding to the image slip. We also average over the three trials.

We can obtain a single quantitative measure of smooth pursuit behavior using the mean squared error (MSE) between greedy action taken by learned policy and the action of the ideal policy that would zero out retinal slip in one time step. The MSE is obtained by averaging over three dimensions. First, we obtain an overall measure by averaging over the 81 different retinal slip conditions shown in Figure 2. Second, we reduce the effect of input variability by averaging over the squared error for 50 actions taken for different pairs of current and past foveal images. Finally, we reduce the effect of stochastic variability introduced by differences in the random initial conditions and choices of target velocities by averaging over the three trials.

Figure 3(a) shows the evolution of the MSE during training for the softmax and Gaussian policies evaluated at 200 equally spaced points along the training. In both cases, we observe a gradual reduction in MSE as the smooth pursuit behavior emerges. The initial MSE of the softmax policy is much larger than the initial MSE of the Gaussian policy. Because we initialize the weights of the Gaussian policy with small random, the mean action will be close to zero for almost all images, and most of the actions chosen will be near zero. On the other hand, for the softmax policy the network outputs with maximum value vary randomly over the entire discrete action range. In addition, the final MSE of the soft max policy is also larger than that of the Gaussian policy. The movie provided in supplemental material traces the evolution of both basis and policy as they evolve simultaneously. Both basis functions and neural network weights are initialized from random.

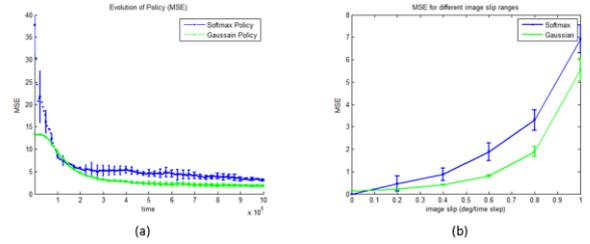

Figure 3(a) Evolution of the mean squared error (MSE) between the learned policy and the "ideal" policy during training. (b) MSE for inputs with different magnitudes of image slip. Error bars indicate the standard deviation in the MSE over trials.

Figure 3(b) shows the MSE of the final policy evaluated for image slips with different magnitudes. The Gaussian policy shows both more accurate (lower MSE) and more precise (lower standard deviation). The advantage becomes more significant as the retinal slip magnitude increases.

Thus, while both policies result in the emergence of smooth pursuit, the Gaussian policy has several advantages over the softmax policy. In addition to the performance advantages noted above, the Gaussian policy also (1) requires fewer parameters: $(N+2)H$ versus $2NK$ where N=300 is the number of basis functions, $H$=5 is the number of hidden units and $K$=11 is the number of discrete actions, (2) does not require us to predefine the possible set and range of actions, which may bias the results of learning, and (3) can give continuous rather than discrete actions.

Finally, we examine the characteristics of the motion tuning of the developing bases. Figure 4 shows three representative bases and their tuning curves obtained from Gaussian policy experiment. We observe that the bases exhibit spatio-temporally profiles very similar to the spatio-temporal Gabor functions used to model the spatio-temporal receptive fields of motion selective visual cortical neurons. This is true in general over all bases (not shown due to space constraints.) We fitted the learned bases using 2D spatial Gabor functions for current and previous frames, assuming the same parameters except for a phase shift, which determines the velocity tuning. The MSE of the residual was around 0.06, indicating excellent fits (recall that bases have unit norm).

We calculated direction and velocity tuning curves for the bases by computing their squared correlation with moving cosine grating with different orientations, directions and velocities. For the direction tuning curves, gratings at the optimal spatio-temporal frequencies were used. For the velocity tuning curves, gratings at the optimal orientation and spatial frequency were used. The first two bases are tuned to similar velocities, but different directions. The third basis is tuned to velocities near zero, and shows little direction selectivity.

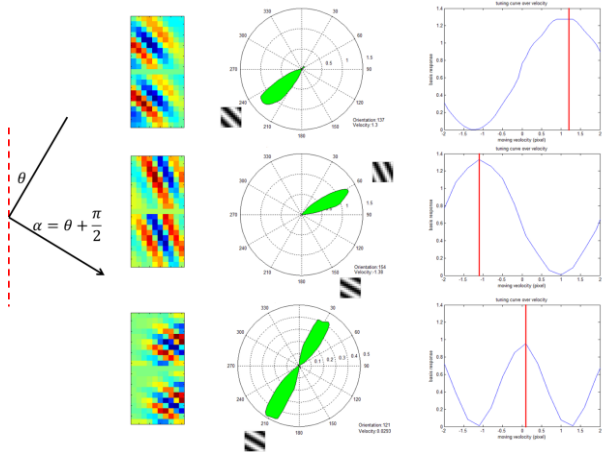

Figure 5 Three representative learned basis functions. The second column shows the spatiotemporal bases as 20 by 10 images, where the top half corresponds to the patch input from time *t*-1 and the bottom to time *t*. The third column shows the directional tuning curve for each basis in polar coordinates where radius indicates response magnitude and angle indicates input velocity direction. The fourth column shows the velocity tuning curve (blue) for each basis. The red line indicates the peak tuning predicted by the fitted Gabor parameters.

Figure 5 shows the histograms of preferred orientations and velocities computed over all bases whose Gabor fit error was smaller than a threshold (0.3). Preferred orientations and velocities were computed based on the fitted Gabor parameters. The peak tuning velocity is given by $2\pi\Delta\varphi/\lambda$, where $\Delta\varphi$ is the phase difference between Gabor fits at time *t* and *t*-1 and $\lambda$ is the spatial wavelength. The distribution of orientations is close to uniform, and most of the bases are tuned to velocities close to zero. This velocity tuning distribution is a consequence of the joint development of perception and behavior. If the eye were stationary, the statistics of retinal slip velocities would be the same as the statistics of the target velocities (evenly distributed between -0.8 and 0.8 deg/s). The efficient encoding hypothesis without our extension to include behavior would predict a more uniform distribution of velocity tunings. The movie provided in the supplemental material gives concrete evidence. At the beginning of the movie when smooth pursuit has not been learned, the neurons display a wider range of speed tuning than in later in training, where they are tuned to smaller velocities.

## IV. DISCUSSION

We have presented an intrinsically motivated model for the co-development of sensory processing and behavior, and have demonstrated that the application of this model in a model of an active eye results in the emergence of neurons selective to visual motion via mechanisms similar to those hypothesized to be used by visual cortical neurons, as well as smooth pursuit behavior. We emphasize that this development is an emergent phenomenon, governed by the properties of the environment and the agent's interactions with it. Nothing in the model itself is specific to motion perception or smooth pursuit. Both sensory processing and behavior co-develop according to a generic criterion: maximizing the fidelity of the neural representation of the sensory input under limited resource constraints.

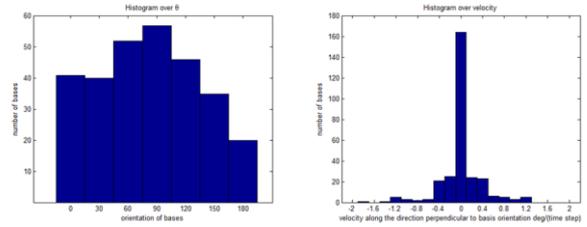

Figure 4 Histograms of the preferred orientations and velocities of the learned bases.

The smooth pursuit model we have presented is consistent with current hypotheses about the neural mechanisms of smooth pursuit. Smooth pursuit eye movements can be divided into two stages, which are divided by a catch-up saccade. In presaccadic pursuit, the eye accelerates to match the velocity of the target. The catch up saccade centers the target in the fovea. In postsaccadic pursuit, velocity matching by smooth pursuit is more accurate. Recent work by Wilmer and Nakayama suggests that pre-saccadic pursuit may be driven by low-level motion-energy-based signals, whereas post-saccadic pursuit may be driven by high-level feature or position based signals [13]. Our model is consistent with the behavior in presaccadic pursuit. The sensory neurons driving the pursuit signal have basis functions and tuning curves similar to the receptive fields and tuning curves of neurons in the primary visual cortex, which are often modeled using motion energy mechanisms (Figure 4).

By stabilizing the image on the retina, smooth pursuit eye movements make the time varying image sequences easier to encode by increasing the correlation between current and delayed visual inputs. For computational convenience, these current and delayed visual inputs have been modeled by a pure delay in this work, but could be implemented in biological neural systems by fast biphasic cells and slow monophasic cells [14]. This stabilization reduces retina slip, resulting in a population of neurons with motion selectivities biased near zero velocity (Figure 5). This result is consistent with Bayesian models of visual motion perception which posit a velocity prior that favors low speeds [15], which have been used to account for psychophysical experiments showing that low contrast stimuli appear to move slower than high contrast stimuli [16][17]. Thus, this model provides a possible explanation for the developmental origins of this velocity prior.

To our knowledge, this model is the first extension of the efficient encoding hypothesis to include the development and effects of behavior. This is a general principle as a model for the development of intrinsically motivated emergent behavior. For example, it can also be used to account for the joint development of stereo disparity tuned neurons and vergence behavior [18]. There are three main contributions of this work. (1) The same framework just operating on different inputs and with different output actions, can result in a different behavior. This demonstrates the generality of our new learning principle. (2) The action space can be more complex (1D vergence versus 2D smooth pursuit). (3) The results are robust to the specific choice action network.

One of the most promising avenues for future development of this model is that it may eventually provide a unified account for the development of a wide range of eye movement behaviors. Past work has often considered different types of eye movement, including pursuit, saccades and vergence separately. However, recent studies are beginning to indicate that these systems share extensive amounts of neural circuitry in their implementation. For example, the majority of pursuit neurons in the frontal eye field (FEF) discharge not only for frontal pursuit but also for vergence eye movements [19]. It has also been demonstrated that pursuit and saccades are not controlled entirely by different neural pathways but rather by similar networks of cortical and subcortical regions, and even by the same neurons in some cases [20]. Future work is necessary to better understand the neural basis of the joint development of sensory and motor learning and its potential relation to the proposed intrinsically motivated learning framework for efficient coding in active perception.

In addition to the hope for a better understanding of the joint development of visual perception and behavior in humans, we believe that our model is also of interest in the current search for artificial systems such as robots that autonomously learn and adapt to their environment. In that regard, one essential property of our model is that it is fully self-calibrating. The vergence model of [18] has been successfully implemented on the humanoid robot head iCub and an analysis of its robustness and adaptive properties has been already conducted by Lonini et al [21]. This study showed that the model is capable to adapt to different kind of perturbations such as blur, or eyes misalignments. This property will become a requirement for the future development of autonomous robots capable of open ended learning.

ACKNOWLEDGMENT

We would like to give special thanks to Luca Lonini, Constantin Rothkopf and Celine Teuliere for their helpful discussions.